\documentclass[pdflatex,sn-mathphys-num]{sn-jnl}

\usepackage{graphicx}
\usepackage{amsmath,amssymb,amsfonts}
\usepackage{multirow}
\usepackage{booktabs}
\usepackage{xurl}
\usepackage{hyperref}
\usepackage{textcomp}
\usepackage{svg}

\newcommand{\code}[1]{\path{#1}}
\begin{document}

\title[Agentic extraction of materials process knowledge]{Agentic schema-guided extraction of materials process knowledge from scientific literature}

\author[]{\fnm{Sameer} \sur{Sadruddin}}
\author*[]{\fnm{Jennifer} \sur{D'Souza}}\email{jennifer.dsouza@tib.eu}

\affil[]{\orgname{TIB Leibniz Information Centre for Science and Technology}, 
\city{Hannover}, \country{Germany}}

\abstract{Materials literature contains detailed experimental knowledge, but procedures, chemical entities and measurements remain difficult to aggregate because they are reported in heterogeneous forms and depend on process-specific context. We present SciKGExtract, a schema-guided framework that combines large-language-model extraction with chemical normalization and agent-based evaluation and refinement before knowledge-graph integration. We evaluate the framework on 176 atomic-layer-deposition papers describing zinc oxide (ZnO) and indium--gallium--zinc oxide (IGZO), together with an expert-annotated full-schema subset. PubChem normalization improves exact-match extraction F1 for every tested model. For ZnO, the best F1 increases from 0.591 for direct normalized extraction to 0.805 with agentic refinement, whereas the best IGZO result is 0.344, revealing the greater difficulty of multicomponent supercycle processes. Evaluation against a deeply nested schema containing 65 experimental properties and 155 quantitative measurement nodes further exposes errors in process segmentation and numerical assignment. These results show that chemical canonicalization and targeted agentic verification provide complementary controls for converting complex materials literature into reusable, machine-actionable experimental knowledge.}


\maketitle

\section{Introduction}

Materials research depends on experimental records that connect composition, synthesis conditions, processing history and measured properties, yet much of this information remains difficult to reuse computationally because a single process may be distributed across methods, results, tables and figure descriptions; the same chemical may appear under formulas, systematic names and laboratory abbreviations; and quantitative values are meaningful only when attached to the correct process instance, quantity kind and unit. Scientific communities are therefore investing in FAIR, machine-actionable infrastructures and initiatives, including NOMAD \cite{sbailo2022nomad}, FAIRmat\footnote{\url{https://www.fairmat-nfdi.eu/fairmat/}}, the U.S. Materials Genome Initiative\footnote{\url{https://www.mgi.gov/}}, and NFDI-MatWerk\footnote{\url{https://nfdi-matwerk.de/}}. At the same time, scientific information extraction has progressed from named-entity recognition, relation extraction and coreference resolution toward direct generation of structured records. Shared tasks and systems including ScienceIE, NLPContributionGraph and SciIE established strong evaluation settings for scientific entities and relations \cite{augenstein2017semeval,dsouza-etal-2021-semeval,luan-etal-2018-multi}, automatically generated scholarly knowledge graphs demonstrated the value of extraction at scale \cite{dessi2020ai}, and domain-adapted models such as MatSciBERT improved text mining of materials terminology \cite{gupta2022matscibert}. More recent work has used LLMs for hierarchical scientific information extraction \cite{dagdelen2024structured}, materials-synthesis benchmarking \cite{yanguas2025benchmarking}, reconstruction of synthesis knowledge from materials literature \cite{li2026materialsreconstruction}, and large-scale materials knowledge-graph construction \cite{bai2025frameworkkg}. These developments show that generative extraction can operate at scientifically useful scale, but scalable population of FAIR resources still depends on reliability when a target record contains dozens of nested properties, multiple experimental instances, quantitative measurements and chemically equivalent surface forms.

Parallel to this transition in information extraction, agentic LLM architectures have established a general paradigm for iterative reasoning and self-correction. ReAct \cite{yao2022react} introduced interleaved reasoning and acting, Reflexion \cite{shinn2023reflexion} used natural-language self-critique as a form of verbal reinforcement, and Self-Refine \cite{madaan2023self} showed that a single model can act successively as generator, critic and refiner through iterative feedback--revision cycles; related forms of iterative model interaction have since been studied in code generation \cite{chen2024teaching,olausson2023selfrepair}, mathematical reasoning \cite{lightman2024lets,shao2024deepseekmath} and embodied agents \cite{huang2023inner}. A complementary development is the use of LLMs as evaluators: G-Eval \cite{liu-etal-2023-g-eval} established rubric-based LLM evaluation, subsequent work examined agreement between strong LLM judges and human preferences \cite{zheng2023judging}, and multidimensional evaluation has used explicit rating criteria for correctness, coherence and related qualities \cite{li2024llms}. Their use for schema-constrained scientific information extraction and knowledge-graph population remains comparatively limited, where refinement must improve not merely the plausibility of generated text but the correctness, completeness and structural placement of machine-actionable values. SciKGExtract adopts evaluation as part of generation rather than as terminal scoring: rubric-based judges identify unsupported or missing information and convert those assessments into actionable revision instructions \cite{llm-rubric,d2025yescieval}, yielding an \emph{extract $\rightarrow$ reflect $\rightarrow$ feedback $\rightarrow$ refine} loop that operationalizes iterative refinement for deeply structured scientific records.

Atomic layer deposition (ALD) provides a demanding materials-science test case for this paradigm. ALD is a surface-controlled vapor-phase deposition technique used to fabricate thin films with precise thickness and composition \cite{kessels2025atomic}, and a useful process record may include deposited material, precursor and co-reactant chemistry, reactor configuration, substrate, pulse and purge times, temperature windows, plasma conditions, growth-per-cycle behaviour, film composition, crystallinity and device properties. ZnO is a mature ALD system with many reported chemistries \cite{paskalevas2022,Tynell2014,gao2019}, whereas indium--gallium--zinc oxide (IGZO) is substantially more complex because deposition can combine In$_2$O$_3$, Ga$_2$O$_3$ and ZnO subcycles in supercycles whose ordering and ratios determine composition and electronic properties \cite{Sheng2019,mackus2019,oh2025b}. SciKGExtract operationalizes the agentic paradigm as a controlled, schema-guided workflow rather than a single LLM call: a semantic schema defines the admissible output structure; an extraction agent generates one or more experimental process records; selected chemical values are resolved to PubChem\footnote{\url{https://pubchem.ncbi.nlm.nih.gov/}} identifiers \cite{kim2016pubchem,pubchem2025}; reflection agents evaluate correctness and completeness; a feedback agent converts evaluations into revision instructions; and accepted records can be prepared for knowledge-graph integration. The framework supports direct extraction, self-refinement, multi-judge refinement and evaluator--critic debate \cite{chan2023chateval,kim2024debate}. Recent work has likewise demonstrated the utility of modular LLM agents for computational materials-science tasks \cite{chaudhari2026matsciagent}; here, that modularity is directed specifically at quality control for literature-derived experimental knowledge, with both the scientific source document and the target schema acting as explicit constraints on generation, evaluation and revision.

To evaluate SciKGExtract in a principled way, we frame the study around five questions that reflect the main methodological components of the framework and the requirements of materials-data reuse: chemical canonicalization, agentic verification, the transition from core-property extraction to deeply structured experimental reconstruction, transfer beyond the ALD domain, and practical quality--cost trade-offs. Our study is therefore organized around the following five research questions.
\emph{RQ1:} To what extent does PubChem-based normalization reduce chemical surface-form disagreement and improve exact extraction across models and material systems?
\emph{RQ2:} Which forms of agentic verification improve structured extraction, and does increasing the number or complexity of interacting agents consistently improve the result?
\emph{RQ3:} How does extraction performance change when the task moves from three core ALD properties to reconstruction of a deeply nested experimental process schema, and which errors become dominant?
\emph{RQ4:} How well does the same schema-guided orchestration transfer to scientific information-extraction tasks outside ALD when evaluated for exact structured correctness?
\emph{RQ5:} What quality--cost trade-offs arise across extraction models and agentic workflows, and does higher model cost correspond to higher extraction quality?
We address these questions through a 176-paper AtomicLimits evaluation \cite{ALDDB}, an expert-annotated full-schema experiment, transfer experiments on established scientific information-extraction datasets, and measurements of model and workflow cost. Exact precision, recall and F1 are treated as the primary metrics whenever the output is intended for structured reuse because a chemically, numerically or relationally incorrect value cannot be made correct by textual resemblance alone.

This work makes several contributions. First, we introduce SciKGExtract, a modular agentic workflow for schema-guided scientific knowledge extraction that unifies structured extraction, chemical normalization, rubric-based evaluation and feedback-driven refinement. Second, we instantiate the workflow on ALD literature and systematically compare multiple proprietary and open-weight extraction models across direct, self-refinement, multi-judge and debate configurations, complemented by a separate controlled full-schema model sweep. Third, we present an evaluation of full-schema experimental process extraction in materials science using an expert-curated gold standard for ZnO and IGZO, thereby moving beyond shallow core-entity recovery to process segmentation, nested property assignment and quantitative reconstruction. Fourth, we demonstrate that agentic reflection and feedback can improve extraction quality while also showing that additional judges or more elaborate interaction do not guarantee better results, and we assess transfer beyond ALD together with the practical cost of alternative workflows. The documented SciKGExtract software is publicly available at \url{https://github.com/sciknoworg/scikgextract}, providing the extraction workflow, schemas, normalization components, prompts and evaluation code needed to reproduce and extend the approach.

\section{Results}

\subsection{A deep experimental schema changes the extraction problem}

\begin{figure}[!htb]
\centering
\includegraphics[width=\textwidth]{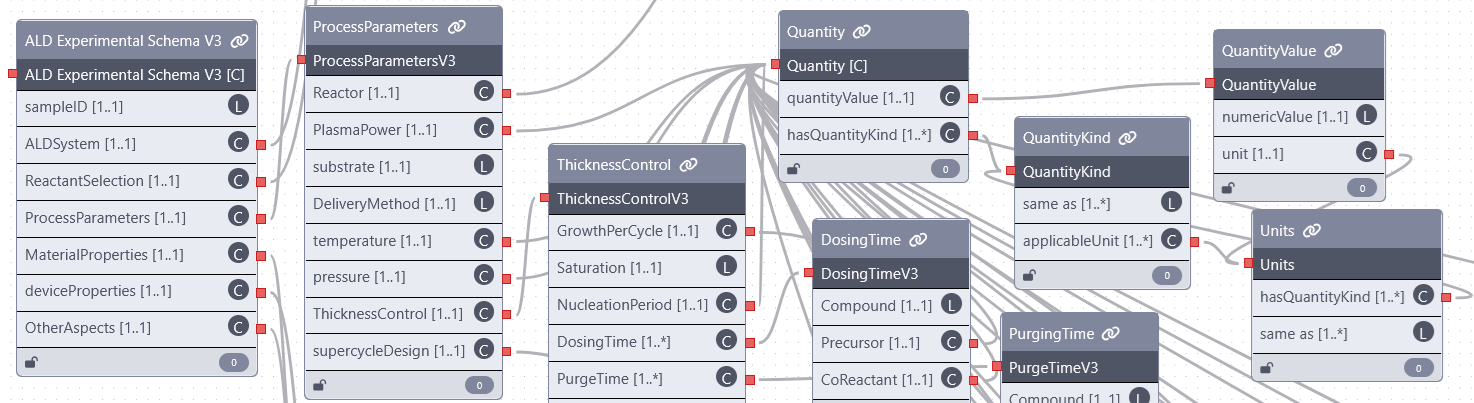}
\caption{\textbf{Partial structure of the ALD experimental process schema.} The schema organizes an ALD process into nested branches for the deposition system, reactants, process parameters, material properties, device properties, and other experimental aspects. Quantitative observations are represented using QUDT-grounded quantity structures. The complete extraction target comprises 65 leaf properties and 155 QUDT-grounded quantity nodes. The complete schema is publicly available as an ORKG template at \url{https://orkg.org/templates/R1434174}.}
\label{fig:ald-schema}
\end{figure}

The extraction target is an expert-validated semantic schema for ALD experimental knowledge \cite{sadruddin2025llms4schemadiscovery,sadruddin2025schema-minerpro}. A partial view is shown in Fig.~\ref{fig:ald-schema}. The root represents one ALD process and connects to nested descriptions of the ALD system, reactant selection, process parameters, material properties, device properties and additional experimental aspects. In its programmatic form the target contains 65 leaf properties. Quantitative values are represented through 155 QUDT-grounded quantity nodes that preserve the quantity kind, numerical value and unit \cite{fairsharing_qudt_2025}. The task therefore requires more than recognizing materials and chemicals: an extractor must identify how many processes are reported, assign observations to the correct process and preserve the relations among operating conditions and measurements.

\begin{figure}[t]
\centering
\includegraphics[width=\textwidth]{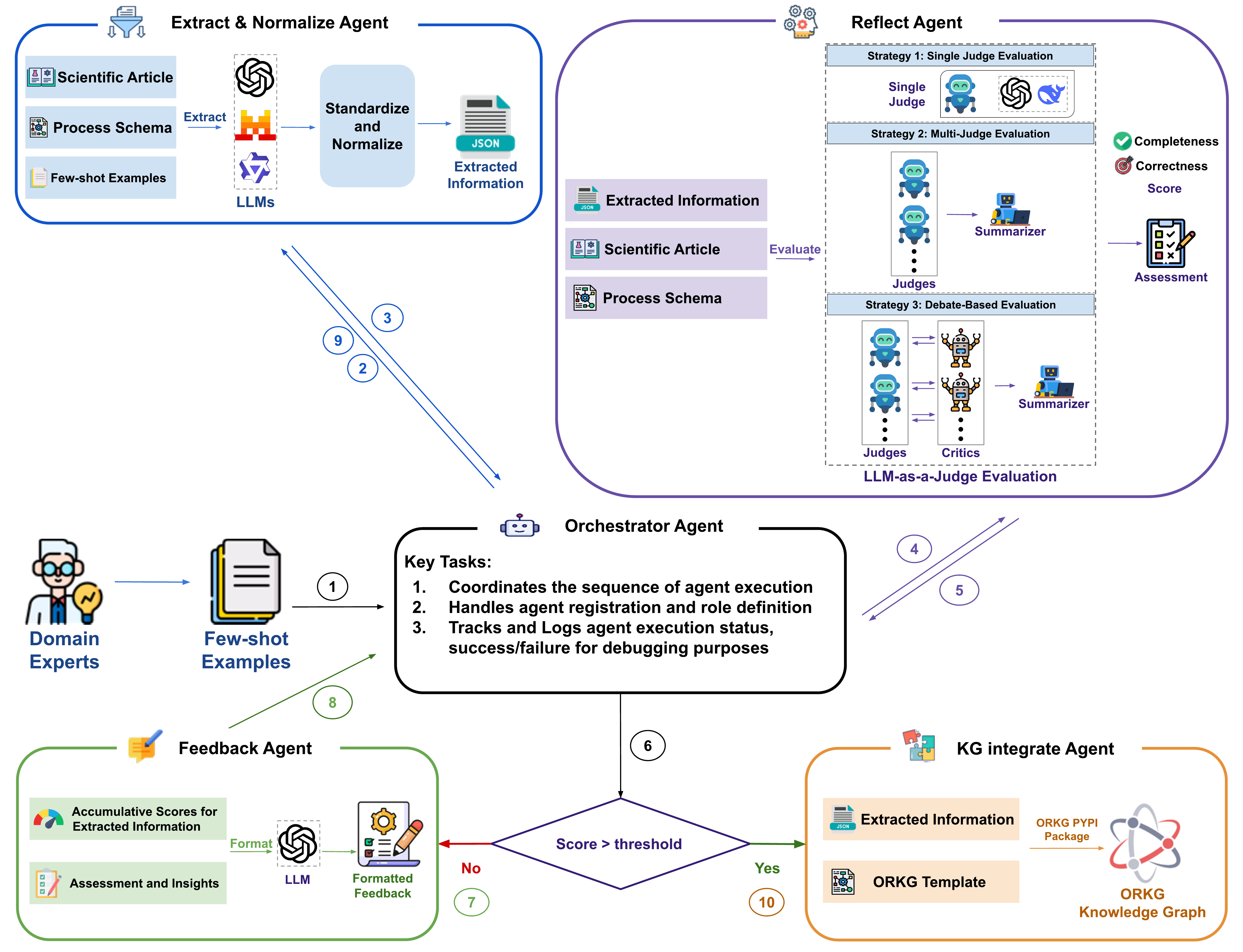}
\caption{\textbf{SciKGExtract orchestration workflow.} The orchestrator coordinates Extract \& Normalize, Reflect, Feedback and KG Integration agents. Structured process records can be evaluated by a single judge, multiple judges or evaluator--critic pairs and revised before accepted records are prepared for knowledge-graph integration.}
\label{fig:workflow}
\end{figure}

SciKGExtract operationalizes this target through the agentic workflow in Fig.~\ref{fig:workflow}. The extraction agent receives the scientific article, target schema and optional examples and produces typed process records. Optional normalization attaches canonical identifiers to selected chemical values. The reflection stage evaluates extracted records against the source article and schema under correctness and completeness rubrics. When the configured acceptance criterion is not met, the feedback agent transforms judge rationales into revision instructions and the extraction is repeated. The same architecture therefore supports controlled comparisons of direct extraction, self-refinement, multiple independent judges and evaluator--critic debate without changing the target representation.

\subsection{Chemical normalization improves exact agreement across models}

To address RQ1, we first evaluate three core ALD properties for which the AtomicLimits database\footnote{\url{https://www.atomiclimits.com/alddatabase/}} provides curated information: deposited material, precursor and co-reactant. The evaluation comprises 162 ZnO and 14 IGZO papers. Table~\ref{tab:best-ald} reports, for each extraction model and material, the best-performing tested refinement strategy according to normalized exact-match F1. PubChem normalization improves F1 for every model in both material systems.

For ZnO, Llama~3.3~70B with debate-based refinement achieves the highest result, with precision 0.780, recall 0.831 and F1 0.805. Among the remaining models, two-judge refinement produces F1 values of 0.697 for GPT-4o, 0.712 for Mistral Large, 0.719 for Gemma~4~26B and 0.673 for Ministral~3B. The normalization gain is particularly large for the smaller Ministral model, whose F1 rises from 0.473 to 0.673 in its best configuration.

IGZO remains much more difficult. The best normalized exact F1 is 0.344 for Llama~3.3~70B with two-judge refinement, followed by 0.334 for Ministral~3B with three judges and 0.302 for Mistral Large with three judges. Normalization nevertheless improves every configuration shown in Table~\ref{tab:best-ald}; for example, the best Ministral~3B configuration increases from 0.162 to 0.334 F1. The persistence of the ZnO--IGZO performance gap after normalization shows that the remaining difficulty cannot be explained only by chemical naming variation.

\begin{table*}[t]
\caption{\textbf{Best exact-match ALD result for each extraction model after PubChem normalization.} Unnormalized F1 is shown for the same refinement configuration to isolate the effect of normalization. MJ2 and MJ3 denote multi-judge refinement with two and three judges, respectively.}
\label{tab:best-ald}
\centering
\small
\begin{tabular}{lllrrrrr}
\toprule
\textbf{Material} & \textbf{Model} & \textbf{Strategy} & \textbf{Unnorm. F1} & \textbf{Precision} & \textbf{Recall} & \textbf{Norm. F1} & \textbf{Accuracy}\\
\midrule
ZnO & GPT-4o & MJ2 & 0.591 & 0.658 & 0.741 & 0.697 & 0.83\\
ZnO & Llama 3.3 70B & Debate & 0.761 & 0.780 & 0.831 & \textbf{0.805} & 0.91\\
ZnO & Mistral Large & MJ2 & 0.629 & 0.696 & 0.729 & 0.712 & 0.89\\
ZnO & Gemma 4 26B & MJ2 & 0.582 & 0.705 & 0.734 & 0.719 & 0.90\\
ZnO & Ministral 3B & MJ2 & 0.473 & 0.637 & 0.713 & 0.673 & 0.83\\
\midrule
IGZO & GPT-4o & MJ2 & 0.202 & 0.239 & 0.304 & 0.268 & 0.82\\
IGZO & Llama 3.3 70B & MJ2 & 0.253 & 0.298 & 0.406 & \textbf{0.344} & 0.83\\
IGZO & Mistral Large & MJ3 & 0.187 & 0.272 & 0.339 & 0.302 & 0.82\\
IGZO & Gemma 4 26B & MJ3 & 0.164 & 0.272 & 0.269 & 0.271 & 0.80\\
IGZO & Ministral 3B & MJ3 & 0.162 & 0.324 & 0.345 & 0.334 & 0.81\\
\bottomrule
\end{tabular}
\end{table*}

\begin{figure}[t]
\centering
\includegraphics[width=\textwidth]{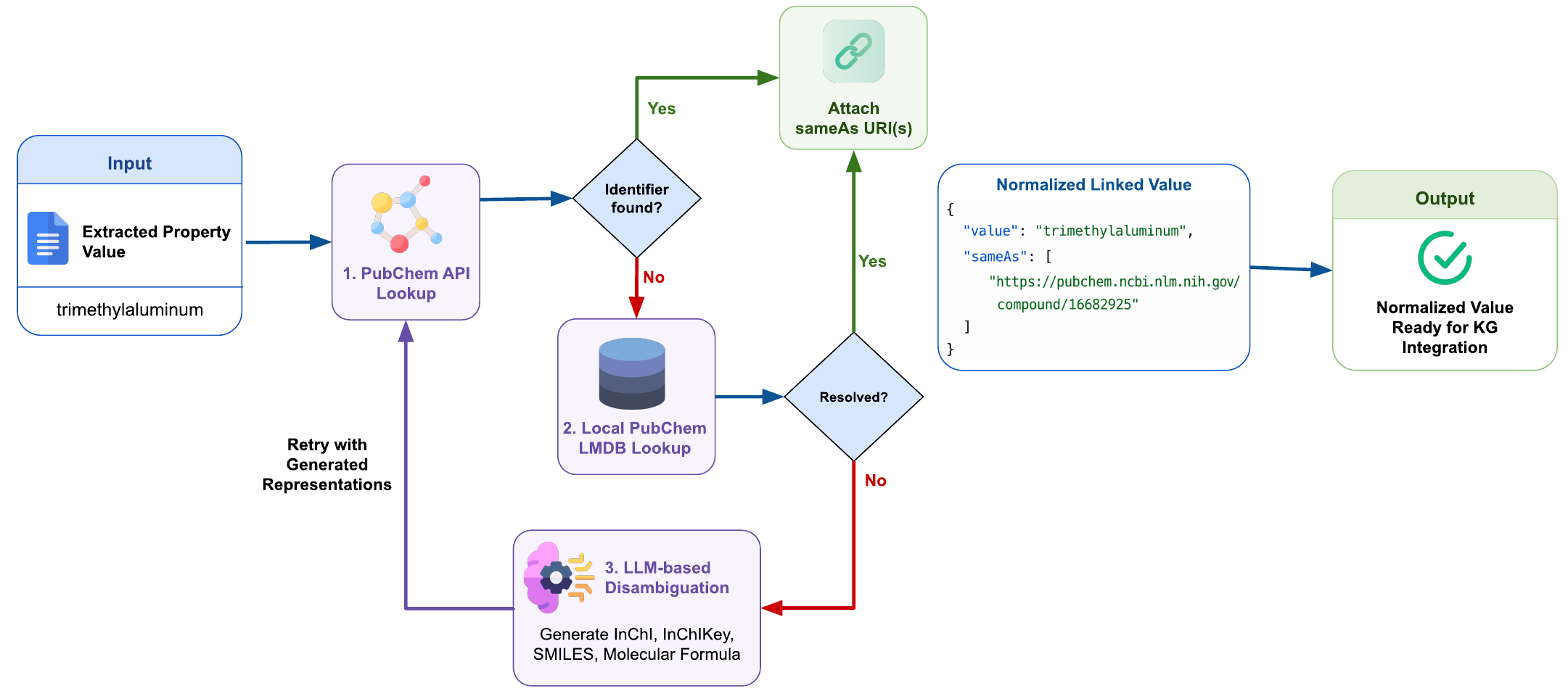}
\caption{\textbf{PubChem-based normalization in SciKGExtract.} Extracted chemical values are resolved through API and local lookup components, with an optional language-model disambiguation step for unresolved forms. The original extracted value is preserved and linked to resolved PubChem identifiers rather than replaced.}
\label{fig:normalization}
\end{figure}

The normalization mechanism is shown in Fig.~\ref{fig:normalization}. SciKGExtract preserves the extracted surface form but attempts to enrich it with PubChem identifiers \cite{pubchem2025}. The lookup first uses PubChem services and a local synonym index; unresolved forms can be reformulated into alternative chemical representations and queried again. This representation allows abbreviations, formulas and systematic names to converge on a common identifier while retaining the original text for provenance.

\subsection{Agentic refinement helps, but more agents are not always better}

To address RQ2 and separate the effect of workflow design from model identity, we evaluate Llama~3.3~70B across the complete progression from direct extraction to self-refinement, multi-judge refinement and debate (Table~\ref{tab:llama-ablation}). For ZnO, normalized exact-match F1 increases from 0.591 for direct extraction to 0.617 with self-refinement, 0.724 with two independent judges and 0.805 with debate. The 0.214 absolute gain from direct extraction to debate is accompanied by higher precision and recall, indicating that refinement is not simply increasing the number of extracted values.

IGZO behaves differently. Direct extraction reaches 0.277 F1 and self-refinement reaches 0.298, while two-judge refinement gives the best score of 0.344. Debate with GPT-5 as both evaluator and critic reaches 0.306; replacing the critic with DeepSeek R1 increases this to 0.331 but still does not surpass the simpler two-judge workflow. The same non-monotonic pattern occurs when moving from two to three judges for several extraction models. Additional agent interaction is therefore useful only when the resulting feedback corrects errors that matter for the target metric; it can also redirect an otherwise adequate extraction or optimize a judgment criterion that is imperfectly aligned with exact gold-standard matching.

\begin{table*}[t]
\caption{\textbf{Effect of refinement strategy for Llama 3.3 70B after chemical normalization.} Debate-self uses GPT-5 as evaluator and critic; debate-cross uses GPT-5 as evaluator and DeepSeek R1 as critic.}
\label{tab:llama-ablation}
\centering
\small
\begin{tabular}{llrrrr}
\toprule
\textbf{Material} & \textbf{Strategy} & \textbf{Precision} & \textbf{Recall} & \textbf{F1} & \textbf{Accuracy}\\
\midrule
ZnO & Direct & 0.577 & 0.607 & 0.591 & 0.88\\
ZnO & Self-refinement & 0.599 & 0.636 & 0.617 & 0.89\\
ZnO & Multi-judge, two judges & 0.702 & 0.748 & 0.724 & 0.87\\
ZnO & Debate & 0.780 & 0.831 & \textbf{0.805} & 0.91\\
\midrule
IGZO & Direct & 0.244 & 0.319 & 0.277 & 0.82\\
IGZO & Self-refinement & 0.261 & 0.348 & 0.298 & 0.83\\
IGZO & Multi-judge, two judges & 0.298 & 0.406 & \textbf{0.344} & 0.83\\
IGZO & Debate-self & 0.271 & 0.351 & 0.306 & 0.81\\
IGZO & Debate-cross & 0.305 & 0.362 & 0.331 & 0.85\\
\bottomrule
\end{tabular}
\end{table*}

The ZnO--IGZO contrast reflects a materials-specific increase in structural complexity rather than only a difference in entity vocabulary. A ZnO process is commonly anchored by one deposited material and a relatively regular precursor/co-reactant pair. IGZO descriptions can contain several oxide subcycles, supercycle ratios, repeated reactants and composition-dependent properties that must be associated with the correct combined process. The decline from 0.805 to 0.344 in the best exact F1 therefore quantifies the difficulty introduced by multicomponent process structure even after chemical canonicalization and agentic checking.

\subsection{Expert evaluation exposes the difficulty of full experimental reconstruction}

To address RQ3, we move beyond the three AtomicLimits fields available at scale and conduct a second, stricter experiment against expert annotations of the complete ALD schema. A domain expert annotated eight thermal-ALD ZnO process instances and five plasma-enhanced-ALD IGZO process instances from two representative papers, covering reactor conditions, reactant parameters, timing, growth characteristics, film composition and device metrics. Because many values are numerical and must be assigned to the correct process and property, the evaluation uses direct precision, recall and F1.

Table~\ref{tab:deep-schema} reports a controlled before/after reflection experiment with GPT-4o, Mistral Large, Qwen3~30B and Qwen3~235B. For ZnO, reflection improves GPT-4o from 0.41 to 0.44 F1, Mistral Large from 0.51 to 0.53 and Qwen3~30B from 0.40 to 0.50; Qwen3~235B remains essentially unchanged. IGZO is again more difficult. Mistral Large improves from 0.20 to 0.32 and Qwen3~30B from 0.41 to 0.46, whereas GPT-4o decreases from 0.48 to 0.45 and Qwen3~235B remains at 0.29. Reflection is therefore a mechanism for correcting errors, not a guarantee of improvement.

\begin{table*}[t]
\caption{\textbf{Exact-match full-schema extraction on expert-annotated ALD processes before and after one reflection--feedback cycle.} This experiment uses the complete 65-leaf-property target and a separate controlled model sweep from the 176-paper refinement comparison.}
\label{tab:deep-schema}
\centering
\small
\begin{tabular}{llrrrrrr}
\toprule
 & & \multicolumn{3}{c}{\textbf{ZnO}} & \multicolumn{3}{c}{\textbf{IGZO}}\\
\cmidrule(lr){3-5}\cmidrule(lr){6-8}
\textbf{Model} & \textbf{Setting} & \textbf{P} & \textbf{R} & \textbf{F1} & \textbf{P} & \textbf{R} & \textbf{F1}\\
\midrule
GPT-4o & Before & 0.42 & 0.39 & 0.41 & 0.50 & 0.49 & 0.48\\
GPT-4o & After & 0.45 & 0.42 & 0.44 & 0.48 & 0.45 & 0.45\\
Mistral Large & Before & 0.53 & 0.50 & 0.51 & 0.24 & 0.18 & 0.20\\
Mistral Large & After & 0.54 & 0.51 & \textbf{0.53} & 0.37 & 0.30 & 0.32\\
Qwen3 30B & Before & 0.42 & 0.39 & 0.40 & 0.43 & 0.42 & 0.41\\
Qwen3 30B & After & 0.52 & 0.49 & 0.50 & 0.50 & 0.47 & \textbf{0.46}\\
Qwen3 235B & Before & 0.50 & 0.47 & 0.49 & 0.33 & 0.30 & 0.29\\
Qwen3 235B & After & 0.50 & 0.46 & 0.48 & 0.33 & 0.30 & 0.29\\
\bottomrule
\end{tabular}
\end{table*}

Coverage analysis shows why the full-schema task is harder than the core-property benchmark. The expert IGZO annotations contain on average 61.8 populated properties per process, whereas one-shot model outputs contain approximately 75, indicating over-generation rather than only omission. For ZnO, the expert average is 35.5 properties compared with approximately 41 in one-shot extraction. Models also differ in process segmentation: the expert annotation contains eight ZnO process instances, whereas the tested one-shot systems recover different subsets of those processes. These results shift the dominant error mode from isolated entity recognition to process segmentation, property assignment and control of unsupported detail.

\subsection{Transfer experiments test portability beyond ALD}

To address RQ4, we exploit the fact that SciKGExtract is driven by an external schema rather than ALD-specific code and test the same orchestration on the biomedical BioRED and BC5CDR benchmarks \cite{luo2022biored,li2016biocreative}. These experiments are used as portability tests rather than direct state-of-the-art comparisons because training regimes, task formulations and available supervision differ across the literature. Table~\ref{tab:transfer} reports the best exact-match configuration observed for each benchmark. On BioRED, Claude Sonnet~4.6 with cross-model debate reaches precision 0.3145, recall 0.2580 and exact F1 0.2834; on BC5CDR, Claude Sonnet~4.6 with two-judge refinement reaches precision 0.6985, recall 0.7542 and exact F1 0.7253. The difference between the two datasets shows that transferring the orchestration does not imply uniform task difficulty or uniform gains from the same refinement strategy, but the framework itself can be reused by changing the external schema and typed output model rather than rewriting ALD-specific extraction logic.

\begin{table}[t]
\caption{\textbf{Best exact-match transfer result observed on each biomedical benchmark.}}
\label{tab:transfer}
\centering
\small
\begin{tabular}{lllrrr}
\toprule
\textbf{Dataset} & \textbf{Model} & \textbf{Strategy} & \textbf{Precision} & \textbf{Recall} & \textbf{Exact F1}\\
\midrule
BioRED & Claude Sonnet 4.6 & Debate-cross & 0.3145 & 0.2580 & 0.2834\\
BC5CDR & Claude Sonnet 4.6 & Multi-judge, two judges & 0.6985 & 0.7542 & 0.7253\\
\bottomrule
\end{tabular}
\end{table}

\subsection{Extraction quality is not proportional to model cost}

To address RQ5, the experiment log records provider cost and token usage for every model request. Average extractor request cost in the executed runs ranges from approximately \$0.00061 for Gemma~4~26B and \$0.00139 for Llama~3.3~70B to \$0.017 for GPT-4o and \$0.020 for Claude Sonnet~4.6. Mistral Large averages approximately \$0.00424 per request and Ministral~3B approximately \$0.00373, the latter reflecting substantially longer prompts and completions in the logged runs. These measurements depend on provider pricing and token usage at the time of execution and are not permanent model characteristics.

The quality ranking is nevertheless clearly not a price ranking. Llama~3.3~70B produces the best normalized ZnO result while remaining among the lowest-cost extraction models in the experiment. Conversely, judge-heavy workflows add several model calls and can become substantially more expensive without guaranteed gains, as illustrated by IGZO. For large-scale materials extraction, the workflow should therefore be optimized as a system: a relatively inexpensive extractor combined with canonicalization and selective verification can be preferable to uniformly invoking the most expensive model.

\section{Discussion}

The experiments support a view of literature extraction as a materials data-engineering problem rather than a single-model prediction task. The schema, normalization and reflection components address distinct error sources. The semantic schema specifies what can be represented and exposes process-level structure. Chemical normalization collapses surface-form variation that would otherwise fragment equivalent materials and reactants. Agentic reflection targets unsupported, missing or structurally inconsistent content after generation. The consistent normalization gains and the more variable reflection gains show that these mechanisms are complementary rather than interchangeable.

The contribution is complementary to recent materials-language-model systems. Large-scale studies have shown that LLM-based extraction can reconstruct synthesis knowledge and populate materials knowledge graphs at substantial scale \cite{li2026materialsreconstruction,bai2025frameworkkg}. Modular materials agents have also demonstrated that specialized LLM components can coordinate retrieval, simulation and structure-oriented tasks \cite{chaudhari2026matsciagent}. SciKGExtract addresses a different bottleneck: how to quality-control a deeply structured experimental record when the schema is fixed in advance and correct population depends on process segmentation, quantitative property assignment and chemical canonicalization. The experiments therefore test not only whether an LLM can generate plausible structured text, but whether the resulting records remain usable as interoperable materials knowledge.

Chemical canonicalization is particularly important for materials knowledge bases because cross-paper reuse depends on identity, not only textual similarity. The core ALD results show that a model can identify the correct chemistry conceptually while still fail exact matching because the paper and gold standard use different names or formulas. Resolving those forms to PubChem identifiers substantially improves agreement across every tested extraction model. This suggests that identifier linking should be integrated into extraction workflows rather than postponed to a separate database-cleaning stage.

Agentic verification has a different profile. For ZnO, Llama~3.3~70B improves steadily from direct extraction through self-refinement, multi-judge refinement and debate. For IGZO, the best result is obtained with the simpler two-judge workflow, while additional debate does not improve F1. The full-schema experiment is similarly mixed: several models improve after one reflection cycle, one remains unchanged and one loses accuracy on IGZO. These findings qualify a central assumption of iterative-refinement paradigms: feedback can improve a candidate output, but additional agents or additional interaction are not themselves evidence of greater reliability. In schema-constrained scientific extraction, the value of reflection depends on whether an evaluator identifies a correctable error and produces feedback that is aligned with exact structured correctness, process identity and property assignment.

The ZnO--IGZO gap is scientifically informative because it identifies process complexity as a major determinant of extraction performance. IGZO combines multiple oxide subcycles, supercycle ratios and composition-dependent measurements. Correct extraction therefore requires long-range document interpretation and assignment of properties to the appropriate process instance. Even after chemical normalization, the best exact F1 remains 0.344, providing a stringent measure of the difficulty of multicomponent process reconstruction and a more realistic benchmark for systems intended to populate materials databases.

The expert full-schema evaluation further shows that success on material, precursor and co-reactant fields does not imply successful reconstruction of an experiment. The principal difficulties become segmentation of multiple processes, hierarchical alignment of nested properties, and assignment of dense numerical information. The expert annotation also revealed an additional ZnO process only after visual information was considered, emphasizing that complete materials extraction is inherently multimodal. The present study is text-centred; integrating tables, plots, schematics and image-derived evidence is a necessary next step.

The study has several limitations. The deep-schema gold standard covers two representative papers and thirteen text-derived process instances, although each instance contains many property--value decisions. Expanding expert annotation across more materials and deposition chemistries would support stronger variance estimates and property-level error analysis. The transfer experiments demonstrate portability of the orchestration but are not intended as state-of-the-art benchmark claims. Model APIs, model versions and provider pricing also change rapidly, so the cost analysis should be interpreted as a measurement of the executed configurations rather than a permanent ranking.

Despite these limitations, the results establish practical principles for materials-literature extraction. Exact structured evaluation should be the primary basis for assessing outputs intended for machine-actionable reuse, canonical identifiers should be incorporated into the extraction pipeline, and deep-schema evaluation should measure process identity and property assignment rather than entity presence alone. Embedding-based similarity measures can still be useful as diagnostics, but they require careful implementation and interpretation for numerical and identifier-bearing records because proximity in an embedding space does not establish scientific equivalence between quantities, identifiers or process assignments. Agentic verification should likewise be used selectively and evaluated empirically rather than assumed to improve with additional judges. Together, these principles make extracted records more suitable for interoperable materials knowledge bases and provide a foundation for downstream comparison, retrieval and reasoning over experimental literature.

\section{Methods}

\subsection{SciKGExtract architecture}

SciKGExtract is implemented as a modular sequential workflow coordinated by an orchestrator, with the overall control flow summarized in Fig.~\ref{fig:workflow}. Four task-specific components are used: Extract \& Normalize, Reflect, Feedback and KG Integration. The orchestrator maintains the shared state, including the scientific article, target schema, typed output model, model configuration, candidate extraction, normalization status, evaluation results and feedback. The same underlying orchestration is used throughout the experiments; the tested refinement strategies differ primarily in how the evaluation and feedback stages are configured rather than in the extraction target or overall processing pipeline.

The Extract \& Normalize component prompts an LLM to generate one or more structured process instances that conform to the supplied schema. The schema is treated as authoritative for property names, value types, nesting and cardinality constraints, while optional few-shot examples provide domain-specific demonstrations of how textual evidence should be mapped into that structure. Outputs are parsed through typed data models and can be checked by a JSON-schema validator, after which cleaning removes residual null values and empty structures. When normalization is enabled, selected structured values are enriched with external identifiers before they proceed to subsequent evaluation. This arrangement separates generation, structural validation and canonicalization while keeping them within a single extraction stage.

The Reflect component evaluates each extracted process against both the scientific source and the target schema using correctness and completeness as the primary rubrics. Correctness captures unsupported values, incorrect units and schema inconsistencies, whereas completeness captures omitted process instances or missing information required to reconstruct the reported experiment. Rubric-based evaluation follows the LLM-as-a-judge paradigm \cite{liu-etal-2023-g-eval,d2025yescieval}. Depending on the experimental setting, this stage is instantiated using a single evaluator, multiple independent judges, or an evaluator--critic configuration. The Feedback component converts judge ratings and rationales into revision instructions that are returned to the extractor when refinement is required. The feedback loop shown in Fig.~\ref{fig:workflow} therefore represents an actual execution path in the system: candidate records can be revised and re-evaluated before being accepted rather than merely scored once.

The KG Integration component transforms accepted process records into graph-oriented representations suitable for resources such as the Open Research Knowledge Graph \cite{auer2025open}. This final stage is deliberately separated from extraction and quality control so that only records that have passed the configured validation, normalization and reflection steps are prepared for downstream graph integration. The architecture is configuration-driven, allowing extraction, validation, normalization, cleaning and reflection to be enabled independently and model providers to be changed without altering the target schema.

\subsection{ALD schema and task representation}

The ALD extraction target is derived from an expert-validated semantic model for experimental ALD knowledge \cite{sadruddin2025llms4schemadiscovery,sadruddin2025schema-minerpro}, a partial view of which is shown in Fig.~\ref{fig:ald-schema}. One ALD process is represented as the lead resource and is connected to nested descriptions of the deposition system, reactant selection, process parameters, material properties, device properties and additional experimental information. The complete programmatic target extends substantially beyond the portion visible in the figure and contains 65 leaf properties. Quantitative measurements are represented using QUDT-grounded quantity structures \cite{fairsharing_qudt_2025}, yielding 155 quantity nodes when the schema is expanded.

This hierarchical representation directly defines the extraction task. The model is not asked to generate an unconstrained textual summary of an experiment, but to instantiate one or more process resources whose properties, nesting and value types conform to the predefined schema. Consequently, an extracted value can be textually correct but structurally incorrect if it is attached to the wrong process instance or placed under the wrong schema branch. A deposition temperature, for example, must be associated with the appropriate process-parameter structure, whereas a mobility measurement belongs to the device-properties branch of the corresponding process. The branching visible in Fig.~\ref{fig:ald-schema} therefore reflects part of the structural reasoning required during extraction in addition to recognizing materials, chemicals and numerical values.

For each document, the extraction target consists of one or more process resources that must satisfy both the evidence contained in the source article and the structural constraints encoded by the schema. The schema is supplied to the LLM in JSON form and mirrored by typed data models used to parse and validate generated records. The same representation underlies the expert full-schema experiment, so the difficulty measured there arises from reconstructing a complete hierarchical experimental record rather than recovering an isolated set of entities or relations.

\subsection{ALD literature corpus}

The principal model/refinement comparison uses papers linked from the AtomicLimits ALD Database \cite{ALDDB}. AtomicLimits provides manually curated deposited-material and reactant information compiled from the ALD literature and major review articles \cite{puurunen2005surface,miikkulainen2013crystallinity,mackus2018synthesis,knoops2019status}. We use 162 ZnO papers and 14 IGZO papers, for a total of 176 papers in the core-property evaluation.

ZnO represents a mature binary-oxide deposition system with many reported chemistries. IGZO represents a multicomponent oxide for which deposition descriptions can contain several subcycles and supercycle ratios \cite{Sheng2019,mackus2019,oh2025b}. PDFs are converted to machine-readable text and tables with Docling \cite{auer2024docling} and stored as Markdown before being passed to SciKGExtract.

\subsection{Chemical normalization}

Selected chemical properties are normalized to PubChem \cite{pubchem2025} using the resolution procedure summarized in Fig.~\ref{fig:normalization}. Normalization is applied to extracted chemical values so that chemically equivalent surface forms can be associated with a common external identifier while the representation used in the source article remains available for provenance. This distinction is important in the ALD literature, where the same material, precursor or co-reactant may be expressed using systematic names, molecular formulas, abbreviations or author-specific variants.

The normalizer first standardizes the extracted surface form and queries the available PubChem lookup components. A cache and local synonym index provide efficient resolution and fallback for forms that are not resolved directly. When enabled, an LLM-based disambiguation step can generate an alternative chemical representation, such as a corrected name, molecular formula or InChI, which is then submitted again to the lookup procedure. As indicated by the fallback sequence in Fig.~\ref{fig:normalization}, deterministic lookup is attempted before generative reformulation, so the language model is used only when the available identifier resources do not resolve the original form.

Successfully resolved values retain the original extracted string and are enriched with the corresponding external identifier rather than being replaced by a canonical text label. The normalization step therefore addresses a specific class of evaluation disagreement: two strings may denote the same chemical entity but fail exact matching because they use different surface forms. By resolving such variants before structured evaluation, the workflow separates chemical naming variation from more substantive extraction errors such as selecting the wrong reactant, associating it with the wrong process, or omitting it entirely.

\subsection{Refinement strategies and models}

Four workflow settings are evaluated. Direct extraction performs one schema-guided extraction pass without reflection. Self-refinement uses a single model family for generation, evaluation and feedback. Multi-judge refinement uses independent judge models whose assessments are consolidated before feedback. Debate refinement introduces evaluator--critic pairs in which the critic can challenge an initial evaluation before feedback is produced \cite{chan2023chateval,kim2024debate}.

The 176-paper ALD comparison evaluates GPT-4o, Llama~3.3~70B, Mistral Large, Gemma~4~26B and Ministral~3B as extraction models. In the two-judge setting, GPT-5 and DeepSeek R1 act as judges and GPT-4o summarizes their assessments. The three-judge setting adds Qwen3~30B Thinking. Debate experiments use GPT-5 as evaluator with either GPT-5 or DeepSeek R1 as critic. Claude Sonnet~4.6 is additionally used in the transfer experiments.

The full-schema expert experiment is a separate controlled model sweep using GPT-4o, Mistral Large, Qwen3~30B and Qwen3~235B \cite{qwen3}. Its purpose is to measure reconstruction of the complete 65-leaf-property schema rather than to rank the newer model set. Results from the two experiment families are therefore reported separately.

\subsection{Expert full-schema annotation}

Two representative ALD studies, one describing ZnO and one describing IGZO, were annotated by a PhD researcher with ALD expertise. The text-based gold standard contains eight thermal-ALD ZnO process instances and five plasma-enhanced-ALD IGZO process instances. Annotation covers reactor conditions, reactant parameters, cycle structure, growth characteristics, composition and device properties. The annotation process included iterative guideline refinement and consistency checking. Inspection of figures identified an additional ZnO process; because the experiments reported here are text-based, that image-derived process is not included in the text-only evaluation. The released annotations are available through the SciKGExtract repository.

\subsection{Transfer benchmarks}

BioRED is a document-level biomedical relation-extraction dataset with multiple biomedical entity and relation types \cite{luo2022biored}. BC5CDR contains chemical and disease entities together with document-level chemical--disease relations \cite{li2016biocreative}. For each dataset, the target schema and typed output model are changed while the SciKGExtract orchestration remains the same. The experiments are used to assess framework portability rather than to claim direct state-of-the-art performance against systems trained under different assumptions.

\subsection{Evaluation metrics}

Exact precision, recall and F1 are the primary extraction metrics. A predicted item counts as correct only when it matches the corresponding gold-standard representation under the evaluation normalization rules, and accuracy is additionally reported for the normalized ALD experiments as recorded in the experiment log. We do not use embedding-based similarity measures as the basis for the conclusions reported here. Such measures can be informative as secondary diagnostics, but their implementation requires particular care for numerical values, chemical identifiers and other structured fields: two serialized values can be close in an embedding space without being scientifically equivalent, and numerical conflation can therefore produce apparently high similarity that has no valid interpretation for structured extraction. Exact agreement is consequently used wherever correctness of the recovered entity, relation, identifier or quantity is the relevant criterion.

\subsection{Cost measurements}

The experiment log records per-request cost and input/output token counts under the provider pricing used when the experiments were executed. Reported costs are averages over the logged requests for a model. They characterize quality--cost trade-offs in the tested workflows and should not be interpreted as fixed price estimates because provider pricing and token usage vary over time.

\subsection{Use of generative AI in manuscript preparation}

Generative AI tools were used to support language editing, phrasing and clarity during preparation of the manuscript. The authors reviewed and validated the resulting text and remain responsible for the scientific content.

\backmatter

\section*{Data availability}

The AtomicLimits source dataset is publicly available through the AtomicLimits ALD Database \cite{ALDDB}. The expert gold-standard annotations used for the full-schema evaluation are available in the SciKGExtract repository at \url{https://github.com/sciknoworg/scikgextract/tree/main/data/gold-standard-annotations}.

\section*{Code availability}

SciKGExtract is available as open-source software at \url{https://github.com/sciknoworg/scikgextract}. The archived software release is available at \url{https://doi.org/10.5281/zenodo.19714379}. The repository contains the extraction workflow, normalization components, schemas, prompts, evaluation code and runnable examples.

\bibliography{mybib}

\section*{Acknowledgements}

The authors thank Eleni Poupaki (Eindhoven University of Technology, The Netherlands), Alex Watkins (University of Warwick, United Kingdom), Erwin Kessels (Eindhoven University of Technology, The Netherlands), and Sören Auer (TIB -- Leibniz Information Centre for Science and Technology, Germany), as well as the wider AWASES project team, for constructive discussions. Access to the large language models used in the experiments was supported by the KISSKI AI Service Center.

\section*{Funding}

This work was supported in part by the ``AI-Aware Pathways to Sustainable Semiconductor Process and Manufacturing Technologies (AWASES)'' project, funded by Merck and Intel. The large language model services used in this work were supported by the KISSKI AI Service Center, funded by the German Federal Ministry of Research, Technology and Space (BMFTR; formerly BMBF) under grant number 01IS22093C.

\section*{Author contributions}

S.S. contributed to the methodology, software development, investigation, data curation, formal analysis, validation, visualization, and preparation of the original manuscript draft. J.D. conceived and supervised the study, contributed to the methodology, experimental design, formal analysis, interpretation of the results, preparation of the original manuscript draft, and review and editing of the manuscript. Both authors discussed the results and approved the final manuscript.

\section*{Competing interests}

The authors declare no competing interests.

\end{document}